\documentclass[letterpaper, 10 pt, conference]{ieeeconf}  

\IEEEoverridecommandlockouts                              
\usepackage{graphicx}
\usepackage{amsmath}
\usepackage{amssymb}
\usepackage{xcolor}
\usepackage[ruled,vlined]{algorithm2e}
\usepackage{algorithmic}
\usepackage[colorinlistoftodos]{todonotes}
\usepackage{adjustbox}
\usepackage{mult irow}
\usepackage{siunitx}
\usepackage[outdir=./]{epstopdf}

\SetKwInput{KwInput}{Input}                
\SetKwInput{KwOutput}{Output}              

\newcommand{\ie}{\textit{i}.\textit{e}.}
\newcommand{\eg}{\textit{e}.\textit{g}.}

\title{\LARGE \bf
Robustifying Reinforcement Learning Agents via Action Space Adversarial Training
}

\author{Kai Liang Tan$^{1}$, Yasaman Esfandiari$^{1}$, Xian Yeow Lee$^{1}$, Aakanksha$^{2}$ and Soumik Sarkar$^{*1}$
\thanks{$^{*}$ Corresponding author}
\thanks{$^{1}$Kai Liang Tan, Yasaman Esfandiari, Xian Yeow Lee, and Soumik Sarkar are with Dept. of Mechanical Engineering, Iowa State University, Ames, IA 50011-2030 {\tt\small {(kailiang, yasesf, xylee, soumiks})@iastate.edu}}%
\thanks{$^{2}$Aakanksha is with Dept. of Computer Science and Engineering, ASET, Amity University Uttar Pradesh, Sector 125, Noida, Uttar Pradesh 201313 {\tt\small aakanksha@student.amity.edu}}
}

\begin{document}

\maketitle
\thispagestyle{empty}
\pagestyle{empty}

\begin{abstract}

Adoption of machine learning (ML)-enabled cyber-physical systems (CPS) are becoming prevalent in various sectors of modern society such as transportation, industrial, and power grids. Recent studies in deep reinforcement learning (DRL) have demonstrated its benefits in a large variety of data-driven decisions and control applications. As reliance on ML-enabled systems grows, it is imperative to study the performance of these systems under malicious state and actuator attacks. Traditional control systems employ resilient/fault-tolerant controllers that counter these attacks by correcting the system via error observations. However, in some applications, a resilient controller may not be sufficient to avoid a catastrophic failure. Ideally, a robust approach is more useful in these scenarios where a system is inherently robust (by design) to adversarial attacks. While robust control has a long history of development, robust ML is an emerging research area that has already demonstrated its relevance and urgency. However, the majority of robust ML research has focused on perception tasks and not on decision and control tasks, although the ML (specifically RL) models used for control applications are equally vulnerable to adversarial attacks. In this paper, we show that a well-performing DRL agent that is initially susceptible to action space perturbations (e.g. actuator attacks) can be robustified against similar perturbations through adversarial training.

\end{abstract}

\section{INTRODUCTION}

Data-driven and learning-based methods are increasingly being applied to cyber-physical systems (CPS) with ubiquitous sensing and advancements in data analytics algorithms. Recent studies have demonstrated the feasibility of deep reinforcement learning (DRL) paradigms being applied on CPS as a controller~\cite{lazic2018data, 7434032, tan2019deep}. The success of these RL paradigms are mainly attributed to the advent of deep neural networks (DNN) that act as expressive decision-making policies. Consequently, adversarial attacks on CPS are inevitable as studies reveal the vulnerability of DNN to adversarial attacks, hence compromising the reliability of RL-based controllers. Success of adversarial attacks in breaking DNNs questions their validity, especially in life- and safety-critical applications such as self-driving cars~\cite{sitawarin2018darts}. The threat caused by attacks on DNNs was first detected while white-box attacks (\ie~attacks that are crafted based on a prior knowledge about the model architecture, hyper-parameters, etc.) were being studied~\cite{carlini2017towards,goodfellow2014explaining,biggio2013evasion,szegedy2013intriguing,kurakin2016adversarial}. Since then, it has also been shown that transfer attacks (\ie~attacks that were crafted for one DNN architecture and mounted on a different DNN architecture) are capable of breaking DNNs~\cite{papernot2017practical}.

Researchers have proposed different methods to defend against adversarial attacks. The most popular method amongst them is adversarial training, where the DNN is trained with an adversarially perturbed dataset~\cite{kurakin2016adversarial,madry2017towards,esfandiari2019saddle}. A more stable adversarial training decouples the min-max problem in the robust optimization problem and solves the problem using Danskin's theorem~\cite{danskin1966theory}. This requires finding the worst case perturbation at each training epoch and updating the model parameters using the dataset which has been augmented by the corresponding worst case attacks. Several methods such as the fast gradient sign method (FGSM)~\cite{goodfellow2014explaining}, Carlini-Wagner (CW)~\cite{carlini2017towards}, projected gradient descent (PGD)~\cite{madry2017towards},  tradeoff-inspired Adversarial defense via surrogate-loss minimization (TRADES)~\cite{DBLP:journals/corr/abs-1901-08573}, and Stochastic Saddle-point Dynamical System (SSDS)~\cite{esfandiari2019saddle} approach are a few such examples which were proposed to find the worst case attack.

In contrast to the defense schemes proposed for learning-based methods, other methods rooted in classical control theory have also been developed to counter adversarial perturbations applied on classical controllers~\cite{AA1, AA2}. In control theory, the notions of robust control and resilient control have been extensively studied. While robust controllers attempt to remain stable and perform well under various (bounded) uncertainties~\cite{zhou1996robust}, resilient controllers try to bring back the system to a gracefully degraded operating condition after an adversarial attack~\cite{resilient}. However, similar notions of robustness and resilience have not been studied well for RL-based controllers, except a few recent works~\cite{icml2019, havens2018online}.

In this study, we investigate the possibility of developing DRL-based controllers that are robust against adversarial perturbations (within specific attack budgets) to the action space (e.g., actuators). Specifically, we develop an algorithm that trains a robust DRL agent via action space adversarial training based on our previous work on gradient-based optimization on action space attacks (MAS-attacks)~\cite{lee2019spatiotemporally}. We would also like to highlight that developing a resilient architecture that is able to recover from adversarial perturbations may not be tractable as DRL algorithms are inherently trajectory-driven. This is because these complex nonlinear models tend to diverge significantly when they encounter an undesirable trajectory.  

\section{RELATED WORKS}
While the robustification of DRL agents under state space attacks and mitigation of actuator attacks in CPS have been studied in both control and ML literature, robustification of DRL agents against actuator attacks have been relatively less studied. In this section, we divide our literature review into control-theory based defense schemes, adversarial attack and defense on DNN, and robustification DRL agents.

\subsection{Control-Theory Based Defense}

Methods to mitigate or improve resiliency of classic controllers against actuation attacks have been extensively studied. For example, authors of~\cite{AA1} proposed a distributed attack compensator which has the capability to recover agents that are under attacks by estimating the nominal behaviors of the individual agent in a multi-agent setting. On the other hand, numerous studies have also developed theoretical bounds on a system's ability to recover from adversarial perturbations and designed corresponding solutions such as decoupling state estimates from control ~\cite{AA2} and ~\cite{AA4} combining a filter, perturbation compensator, and performance controller to re-stabilize a system. Additionally, it has been shown that actuation attacks may go undetected if the attacks are deployed at a higher frequency than sensor sampling frequencies~\cite{AA5}, but these attacks can be mitigated with controllers with multi-rate formulations~\cite{AA6}.

\subsection{Adversarial Attacks and Defenses}

After~\cite{szegedy2013intriguing} exposed the vulnerabilities of DNNs to adversarial attacks, a defense strategy was proposed by~\cite{biggio2013evasion}, which defines a regularization term in the classifier. Concurrently,~\cite{goodfellow2014explaining} introduced Fast Gradient Sign Method (FGSM) which was used by~\cite{kurakin2016adversarial} to craft a defense strategy against iterative FGSM attacks. Departing from the notion of white-box attacks,~\cite{papernot2017practical} showed the vulnerability of DNNs to transfer attacks in which no prior knowledge about the model architecture is needed. Before adversarial training was popularized as a defense method~\cite{madry2017towards,shaham2018understanding,goodfellow2014explaining}, researchers proposed several other defense approaches including defining a network robustness metric~\cite{NIPS2016_6339} and using de-noising auto-encoders to form Deep Constructive Networks~\cite{gu2014towards}. After adversarial training gained popularity,~\cite{papernot2016distillation} proposed defensive distillation as another powerful method for defense, although~\cite{carlini2017towards} devised multiple loss functions to produce attacks that could break this defense mechanism. Since DRL algorithms employ the use of DNNs as policy functions, those models are also found vulnerable to the attacks described above~\cite{behzadan2017vulnerability}.

\subsection{Robust Deep Reinforcement Learning}

In the context of training a DRL agent to be robust against state space attacks where the input to the DRL agent is perturbed,~\cite{pattanaik2018robust} and~\cite{mandlekar2017adversarially} demonstrated that a DRL agent that is robust to parameter and environmental variations can be obtained by adversarially training the agent.. In another study,~\cite{havens2018online} proposed a meta-learning framework where a meta-RL agent has access to two sub-policies. The meta-agent learns to switch between a policy that maximizes reward during nominal conditions and another policy that mitigates and copes with adversarially perturbed states by observing advantage estimates of both policies during deployment.~\cite{pinto2017robust} used a different robustifying scheme by formulating a zero-sum min-max optimization problem where the DRL agent is trained in the presence of another DRL agent that adversarially perturbs the system. Similarly,~\cite{icml2019} demonstrated that DRL agents can be robustified against disturbance forces by training the agent with some noise perturbation in a min-max formulation.
 
\section{METHODOLOGY}

\begin{figure}[t]
  \centering
  \includegraphics[width=.7\columnwidth]{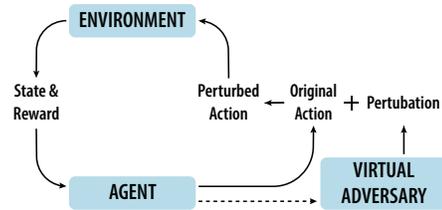}
  \caption{Robustifying DRL agent by perturbing the original agent's action. The perturbation is generated by a white-box adversary, where the adversary has access to the agent's network architecture and parameters.}
  \label{typicalRL}
\end{figure}

We provide a brief overview of DRL algorithms, followed by discussion of robust learning from a robust optimization standpoint. Finally, we formulate the robust RL methodology by combining the RL and robust learning formulation.

\subsection{Reinforcement Learning}\label{rl}
In RL, the goal of an agent is to maximize its cumulative future rewards. A typical setup involves an agent interacting with an environment for a finite number of steps, or until a termination condition is met. Upon termination of an episode, the environment resets to an initial state for the agent and repeats the process again. At each step in the environment $t \in T$, the agent receives a state $s_{t} \in S$ and reward $r_{t} \in R$, then selects an action $a_{t} \in A(s)$ from the agent's policy $\pi$. $T$ denotes the finite number of steps for an environment, $S$ denotes all possible states for an environment, $R$ denotes the cumulative reward for one episode, and $A(s)$ denotes all possible actions conditioned upon the state. Every state observation is quantified with a reward value indicating how valuable that state is for the agent. Specifically, given a finite number of time steps $T$, the agent's goal is to learn an optimal policy which maximizes the cumulative discounted rewards $G_{t}$:

\begin{equation}
    G_{t} = \sum_{t=0}^{T} R = r_{t} + \gamma r_{t+1} + \gamma^{2} r_{t+2} + \cdots + \gamma^{T-1}r_{t}
\end{equation} where gamma $\gamma$ $\in$ $[0, 1)$. A $\gamma$-value of $0$ makes the agent nearsighted (prefer short-horizon rewards), while a value of $1$ makes the agent farsighted (prefer long horizon rewards). Since being in a specific state is a direct result of previous state and action, the agent's policy will evolve over time to refine the understanding of good and bad trajectories $\tau$ (sequence of state-action combination). Ultimately, the goal is to optimize the agent's policy $\pi^{*}(a|s)$ such that the mapping between state and action is optimal.

There are two known methods to optimize a policy, namely action-value and policy gradients. Action-value methods optimize the action value for each state-action pair $Q^{\pi}(s,a)$ as shown in Eq.~\ref{bellman}. Examples of action-value methods include Deep Q-Network (DQN)~\cite{mnih2015human} and DoubleDQN (DDQN)~\cite{van2016deep}. 

\begin{equation}
    Q^{\pi}(s_{t}, a_{t}) = \max_{\pi} \mathbb{E} \big[G_{t} | s_{t}, a_{t}\big]
\label{bellman}
\end{equation} Policy gradient methods optimize a policy parameterized by theta $\theta, \pi_{\theta}(a|s)$, where $\theta$ is directly optimized to maximize the expected reward function $J(\theta)$:

\begin{equation}
    J(\theta) = \sum_{s_{t} \in S} d^{\pi}(s_{t}) \sum_{a_{t} \in A} \pi_{\theta} (a_{t}| s_{t}) Q^{\pi} (s_{t}, a_{t})
\label{policy_gradient}
\end{equation} $d^{\pi}(s_{t})$ denotes the stationary distribution of Markov chain~\cite{hendricks1972stationary} for $\pi_{\theta}$. Examples of policy gradient methods include Trust Region Policy Optimization (TRPO)~\cite{schulman2015trust} and Proximal Policy Optimization (PPO)~\cite{schulman2017proximal}.

\subsection{Robust Optimization Classical Formulation}\label{rotheo}

The robust optimization problem can be defined as~\cite{madry2017towards}:

\begin{figure}[t]
  \centering
  \includegraphics[width=.6\columnwidth,clip,trim={0.0in 0.3in 0.0in 0.8in}]{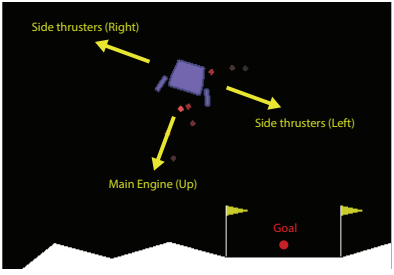}
  \caption{The goal of the agent is to land at the goal. Annotated directional arrows indicate thrust direction of the lunar lander}
  \label{LunarLander}
\end{figure}

\begin{equation}
    {\cal RO}:=\min_{w} \; \mathbb{E}_{(x,y) \sim {\cal D}} \; \Big[\underset{\delta \in {\cal B}}{\max}\;\; L(w,x+\delta,y)\Big]\;
\label{RO}
\end{equation} where $x \in \mathbf{R}^m$ is the dataset under a data distribution ${\cal D}$ with set of labels, $y$. The loss function (\eg~cross-entropy loss) with additive perturbations $\delta\in\mathbf{\cal B}$ is denoted by $L(w,x+\delta,y)$ with $w\in\mathbf{R}^{n}$ as the model parameters (decision variables). Here the saddle point problem is looked at as a composition of an inner maximization and an outer minimization, where the inner maximization tries to find a specific perturbation for each data point $x$ such that the overall loss is maximized. In parallel, the outer minimization aims to achieve the model parameters which minimize the corresponding adversarial loss. 

The next step is to define the attack model.We introduce a specific perturbation $\delta\in\mathbf{\cal B}$ for each data point $x$, where ${\cal B}$ is the set of allowed perturbations (${\cal B} \in\mathbf{R}^m$). This set acts as a normalization for the perturbation power. For example, $l_\infty$ ball around x is a popular way to define the perturbation budget~\cite{goodfellow2014explaining}. There are several attack methods to find the corresponding adversary $\delta$ for each data point $x$ (\ie~FGSM, PGD, etc.). In this paper, we use PGD for finding the adversaries~\cite{madry2017towards} which uses $\alpha$ as the step-size, and can be formulated as:

\begin{equation}
    x_{k+1} = x_{k} + \underbrace{ \alpha \; sgn(\nabla_{x} \big. L(w,x,y))}_\text{$\delta$}
\label{PGD}
\end{equation}

For training a neural network, Stochastic Gradient Descent method is used for solving the outer minimization problem at the maximizer point of the inner problem. This approach is valid as Danskin's theorem proves that the gradients at the inner maximizer act as a valid descent direction for outer loss minimization~\cite{danskin1966theory}.

\subsection{Robust Reinforcement Learning Agents}

To achieve a robust DRL agent, the robust optimization is formulated as a DRL problem. In this paper, we focused on white-box attacks in action-space. As such, the robust optimization problem can be written as:

\begin{equation}
    {\cal RO}:=\max_{\theta} \; \mathbb{E}_{(s,a)} \; \Big[\underset{\delta \in {\cal B}}{\min}\;\; R(s_{t},a_{t}+\delta_{t})\Big]\; 
\label{RORL}
\end{equation} where $(s,a) \in \mathbf{R}$ are state and action pairs. The reward function with additive perturbations $\delta_{t} \in\mathbf{\cal B}$ is denoted by $R(s_{t},a_{t}+\delta_{t})$ where $(s,a)$ are updated based on the policy ($\pi_{\theta}(a|s)$) in each iteration, with $\theta$ as the model parameters.
Note that this formulation is different from the classical ${\cal RO}$ formulation (Eq.~\ref{RO}) in which model parameters are an explicit input to the loss function, whereas in RL the reward function is not an explicit function of model parameters.

The attack formulation in RL is based on our previous work with MAS-attack~\cite{lee2019spatiotemporally}. MAS-attacks are derived from white-box attacks in action space, where the perturbations $\delta_{t}$ for each $(s_{t},a_{t})$ are computed based on complete knowledge of the policy of the agent, $\pi_{\theta}$ (See Fig.~\ref{typicalRL}). Each perturbation $\delta_{t}$ is bounded by $\cal B \in \mathbf{R}$, by projecting the perturbations back into ${\cal B}$. As MAS-attack uses PGD to iteratively find the perturbations, we can formulate the attack as:

\begin{equation}
    a_{k+1}\;=\;a_k-\alpha\;\nabla_{a_{k}} {\cal D}
\label{PGD-RL}
\end{equation} where $k$ is the iteration number, $\alpha$ is the step size and ${\cal D}$ is the action distribution obtained from the agent's policy $\pi_{\theta}$. Here, we note that while the robust optimization formulation was formulated using the reward function, in reality, the reward function is unknown to the DRL agent or the virtual adversary. Hence, the reward function is approximated by the reward-maximizing action distribution ${\cal D}$ for gradient computations~\cite{lee2019spatiotemporally}. After obtaining the adversarial perturbations and adding it to the nominal actions, we train the DRL agent using standard policy gradient methods, which solves the outer maximization problem in the formulation above. Another distinction we would like to note is that while a $k$-step PGD is usually used to compute the adversarial attacks, with $k$ being fixed, we do not adhere to that procedure. Instead, we define a tolerance $\epsilon$ and keep iterating through the PGD process until the adversarial actions saturate. This ensures that the computed adversarial action actually corresponds to bad action rather than being approximated by a fixed $k$-number of gradient steps. The adversarial training algorithm to robustify the DRL agent is shown in Alg.~\ref{alg:MAS}.

\begin{algorithm}
\caption{MAS-Adversarial Training}
\label{alg:MAS}
\SetAlgoLined
\KwInput{episodes $N$, episodic limit $T$, step size $\alpha$, convergence criteria $\epsilon$, budget $B$}
Initialize state $s$, policy $\pi_{\theta}$ \\   
\For{$n\in\{1,...,N\}$}
    {
    \For{$t \in T$}
        {
        Get action distribution $D$ from $\pi_{\theta}(s_t)$ \\
        Sample $a_t$ from $D$ \\
        $a_k = a_t$ \\
        Sample $a_{k+1}$ from $D$\\
        \While{${a}_{k+1} - {a}_{k} \geq \epsilon$}
            {
            ${a}_{k+1} = {a_k} - \alpha \ \nabla_{a_k} \big. D$ \\
            }
        $\hat{\delta_t} = P_{\mathcal{B}}({a}_{k+1} - a_t$) \\
        $\hat{a_t} = a_t + \hat{\delta_t}$ \\
        Step through environment with $\hat{a_t}$
        }
        \If{Time to update}
            {
            Update agent's policy $\pi_{\theta}$
            }
    }
\end{algorithm}

\section{EXPERIMENTS}
\subsection{Environment Setup}
Experiments were conducted in OpenAI gym's Lunar Lander environment \cite{gym}, where the goal is to land the lander safely while minimizing thruster usage. The state space of the environment consist of eight continuous values: x-y coordinates of the lander, velocity in x-y components, angle and angular velocity of the lander, and a boolean contact variable for left-right lander legs (\eg~1 for contact, 0 for no contact). The action space available are two continuous vectors [-1,1]. First vector controls the up-down engine, where values within [-1,0] turns off the engine while values within (0,1] maps to 50\% - 100\% of engine power. Second vector controls left-right engines, where [-1,-0.5] and [0.5,1] controls left and right engine respectively.

For this environment, the agent's goal is to maximize reward. Given an arbitrary starting point (as seen in Fig. \ref{LunarLander}), the RL agent has to land on the landing pad without crashing. The agent receives positive rewards (\eg~between 100 to 140) for landing. The agent will incur negative rewards if the lander moves away from the landing pad. Each successive landing leg contact gives 10 rewards each. Firing the up-down engine cost -0.3 rewards each step. The episode is terminated if the lander crashes or is at rest, which gives -100 and 100 rewards respectively. 

\subsection{Deep Reinforcement Learning Training and Parameters}

For this experiment, we trained a PPO agent with an Actor-Critic architecture~\cite{konda2000actor}. In this architecture, both the actor and the critic share the same network of multi-layer perceptrons made up of dense layers. The actor has an additional dense layer which outputs the policy in the form of a Gaussian model and the critic estimates the value of the action chosen by the actor. To encourage exploration, an entropy term is also added to the loss function which consists of the actor's loss and the critic's loss. The loss function is then jointly optimized using Adam optimizer. 
\\

\section{RESULTS AND DISCUSSION}

\begin{figure}[t]
  \centering
  \includegraphics[width=.9\columnwidth,clip,trim={0.0in 0.15in 0.0in 0.0in}]{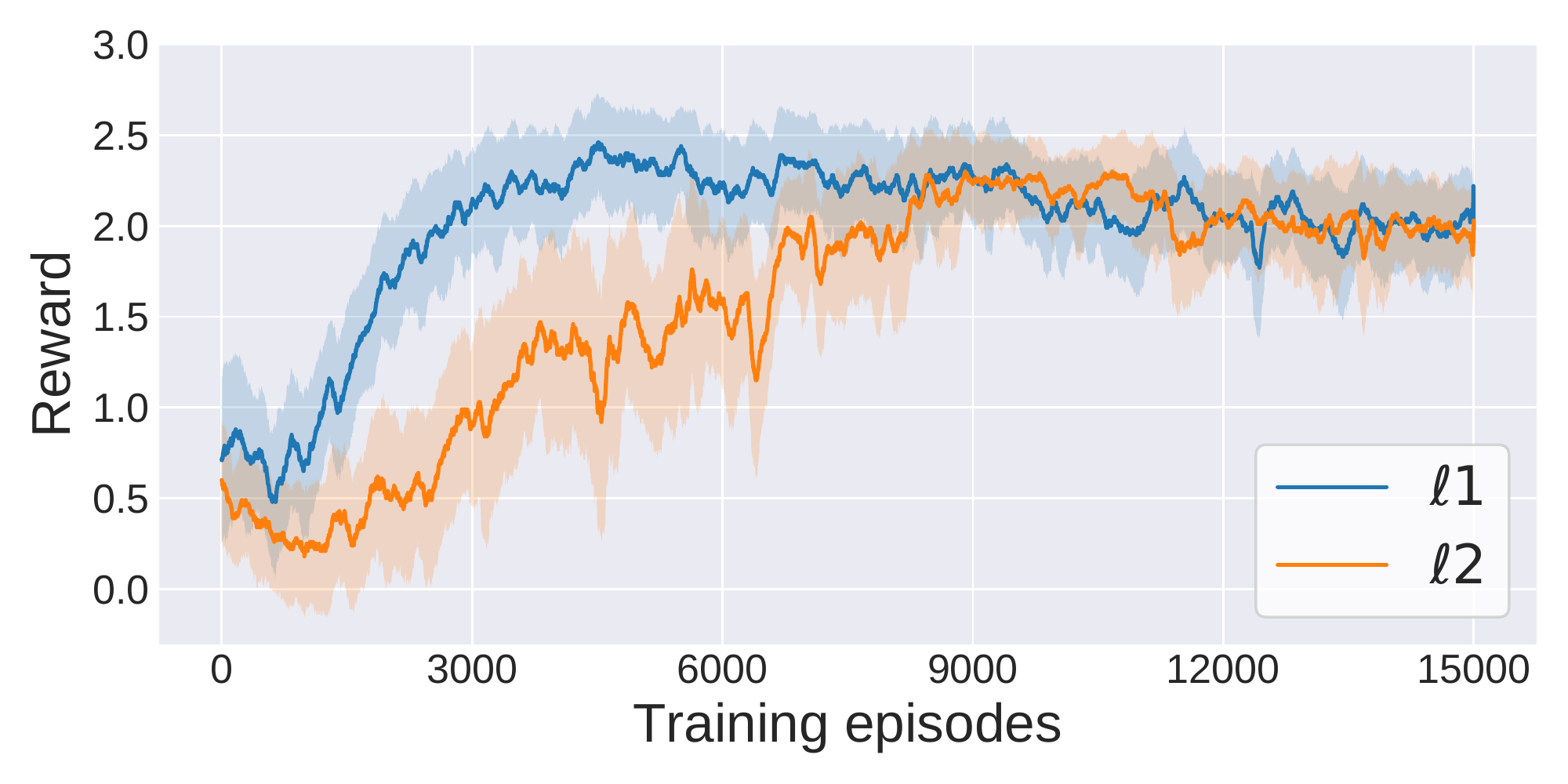}
  \caption{Training plots on robustly trained DRL agents with MAS-attacks ($\ell 1$ and $\ell 2$ projection). Seven agents was trained for each projection method. Each rewards are averaged across all agents, followed by a moving average.}
  \label{training}
\end{figure}

We conducted two different experiments to study the difference in robustifying DRL agents by using $\ell 1$ and $\ell 2$ projection methods with MAS-attack budget of 1 and step size of 3 within Algorithm~\ref{alg:MAS}. 

\subsection{Convergence of Robust Agent}
In this section, we analyzed the convergence plots (Fig.~\ref{training}) of the robust agent empirically. The robust agent was trained with $\ell 1$ and $\ell 2$ perturbations at every step for 15000 episodes. For each projection method, we trained seven DRL agents with the same network architecture and parameters across different seeds. After 15000 episodes, training rewards are averaged across seven agents, followed by a moving window of 100 episodes to obtain results shown in Fig.~\ref{training}. Agents trained with $\ell 1$ converged faster to a higher reward at approximately 4500 episodes as compared to training with $\ell 2$, which stabilized around 9000 episodes. This reveals that it is slightly harder to robustify the agent against $\ell 2$ as compared to $\ell 1$ due to $\ell 2$ attacks being more distributed along action dimensions compared to $\ell 1$.

\begin{center}
    \begin{table}[t]
    \centering
    \caption{Summary statistics of agent's rewards}
    \begin{adjustbox}{width=\columnwidth,center}
    \begin{tabular}{m{0.02\textwidth} c c c c c}
    \hline 
    \hline
    \multirow{3}{*}{\rotatebox[origin=c]{90}{$\boldsymbol{\ell 1}$}} &
    Environment & \multicolumn{2}{c}{Nominal} & \multicolumn{2}{c}{Adversarial} \\
     & Agent & Nominal & Robust & Nominal & Robust \\
    \cline{2-6}
     & Reward & $2.21\pm0.74$ & $2.21\pm0.54$ & $0.74\pm1.33$ & $2.39\pm0.47$\\
    \hline
    \hline
    \multirow{3}{*}{\rotatebox[origin=c]{90}{$\boldsymbol{\ell 2}$}} &
    Environment & \multicolumn{2}{c}{Nominal} & \multicolumn{2}{c}{Adversarial} \\
     & Agent & Nominal & Robust & Nominal & Robust \\
    \cline{2-6}
     & Reward & $2.21\pm0.74$ & $1.74\pm0.63$ & $0.60\pm1.10$ & $2.00\pm0.61$ \\
    \hline
    \hline
    \end{tabular}
    \end{adjustbox}
    \end{table}
    \label{table_statistics}
\end{center}

\begin{figure*}[t]
  \centering
  \includegraphics[width=.9\textwidth,clip,trim={0.0in 0.2in 0.0in 0.0in}]{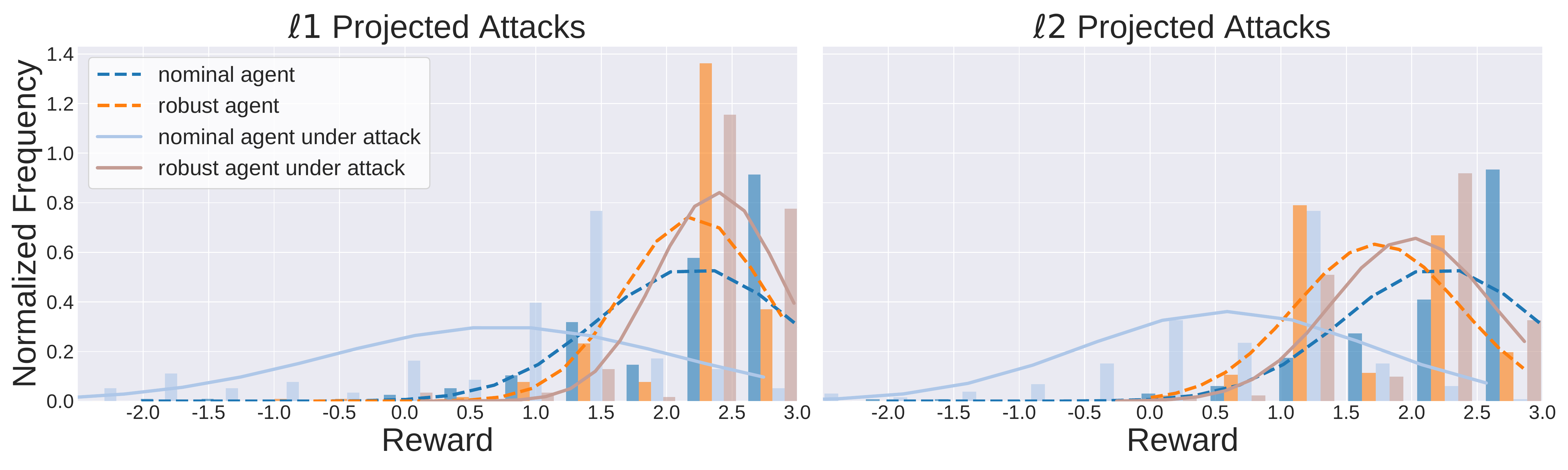}
  \caption{A comparison of reward distributions between a nominally trained agent and adversarially trained agent for both $\ell1$ and $\ell2$ MAS attacks. We observe that the distribution of rewards for a nominal agent shifts to the left and have long tails in the regions with negative rewards when subjected to attacks. In comparison, the distributions of rewards remain similar when the robustly trained agents are subjected to attacks.}
  \label{histogram}
\end{figure*}

\subsection{Performance Comparison}
We compare the performance of the robustly trained DRL agents against the nominally trained DRL agent. Specifically, we tested both DRL agents under two environment settings; nominal and adversarial environment. A nominal environment denotes scenarios where the agent is not attacked, while the agent is attacked in adversarial environments. Fig.~\ref{histogram} shows a histogram of all four possible scenarios for each projection attack. For both projection attacks, the nominal agent tested in the nominal environment is identical. The other three scenarios show slightly different results due to different projection attacks. We summarize the reward distribution of each scenario in Table I. Rewards within the range of 2.5 to 3.0 are successful landing experiments, where the DRL agent learns to land on the landing pad and shuts off it's engines. Rewards between 1.0 to 2.5 are DRL agents that lands successfully but continuously loses rewards as it fails to turn off the engine. Rewards below 1.0 are experiments where the lunar lander crashed or flew out of frame. 

\subsubsection{Nominal environment}
In nominal environments, the nominal agent's performance is identical across both $\ell 1$ and $\ell 2$ plots in Fig.~\ref{histogram} which is expected, as the agent is not attacked. As a result, the agent has a mean reward of $2.21\pm {0.74}$. This indicates the nominal agent successfully lands and turns off its engine. 

Next, we evaluate both the robust agent trained with $\ell 1$ and $\ell 2$ projection attacks in the nominal environment. The rewards for the robust agent in $\ell 1$ and $\ell 2$ are $2.21\pm {0.54}$ and $1.74 \pm{0.63}$ respectively. 
We hypothesize that the nature of $\ell 2$ crafted attacks are more evenly distributed across action dimension, hence it is harder to train and test against $\ell 2$ attacks where else $\ell 1$ distribute attacks into one dimension. These results are counterintuitive to the notion of robustifying a DRL agent, where the expected results of a robust agent should be higher than the nominal agent in the nominal environment. Interestingly, the same behavior has been observed when robustifying DNN used for image classification as demonstrated in~\cite{madry2017towards}.

\subsubsection{Adversarial environment}
In the adversarial environment, the nominal agent's performance in both $\ell 1$ and $\ell 2$ projection attacks dropped significantly as anticipated. The nominally trained agent's policy was trained in environments with no perturbations. The rewards for both $\ell 1$ and $\ell 2$ projection attacks are $0.74 \pm {1.33}$ and $0.60 \pm{1.10}$ respectively. Hence, both projection attacks successfully minimized the nominal agent's reward. In both $\ell 1$ and $\ell 2$ attacks, we observed a high frequency of rewards obtained within the range of 1.0 and 1.5, which corresponds to scenarios where the lander landed but failed to turn off its engine.

For the robust agent, the performance of both $\ell 1$ and $\ell 2$ trained agent increased when compared to the nominally tested robust agent counterpart. The rewards for $\ell 1$ and $\ell 2$ trained agents are $2.39 \pm{ 0.47}$ and $2.00 \pm {0.61}$ respectively. Similar to counter-intuitive observations made earlier, we note that the expected results should be a decrease in rewards compared to the robust agent in the nominal environment. These counter intuitive results reveal an important characteristic of adversarial training defense schemes. We can expect that an agent that has been adversarially trained will perform well when tested in an adversarial environment, but at the cost of a slightly reduced performance when tested in nominal situations. 

Although it is not a direct comparison, it is interesting to note that the robust agent trained with $\ell 1$ projected attacks in the adversarial environment outperforms the nominal agent in the nominal environment. This is likely because the nominal agent can only explore and maximize its reward with familiar trajectories seen during training. For the robust $\ell 1$ agent, the agent's policy explored many more trajectories with the help of adversarial perturbations. Therefore, it has likely found other trajectories with much higher rewards that the nominal agent did not explore.

\section{CONCLUSIONS}
Deep RL based controllers are increasingly popular as they demonstrate a potential for controlling complex CPS. Adversarial attacks on these controllers are emerging, which requires these controllers to be robustified against these attacks. In this work, we formulate the problem of robustifying a DRL agent as a robust optimization problem. We adversarially trained a DRL agent that is subjected to action space perturbations and demonstrate that it still performs robustly in the presence of actuator perturbations. In some cases, it even improved the performance of the agent in the absence of attacks. Hence, we show that it is beneficial to adversarially train a DRL agent. Future direction includes extending this work to different attack models and experimenting with transferability of attacks and defense results.

\section*{ACKNOWLEDGMENT}
This work was supported in part by NSF grant CNS-1845969.

\bibliographystyle{IEEEtran}
\bibliography{IEEEabrv,reference}

\end{document}